\RequirePackage{fix-cm}
\documentclass[letterpaper, 10 pt, conference]{ieeeconf}  

\IEEEoverridecommandlockouts                              

\usepackage{graphicx} 
\usepackage{epsfig} 
\usepackage{times} 
\usepackage{amsmath} 
\usepackage{amssymb}  
\usepackage{booktabs,multirow,makecell}
\usepackage{xcolor}
\usepackage{algpseudocode}
\usepackage{lipsum}
\usepackage{url}
\usepackage{mathtools}
\usepackage{scalerel}
\usepackage{bm}

\let\labelindent\relax
\usepackage{enumitem}

\usepackage{balance}
\usepackage{setspace}
\usepackage[linesnumbered,ruled,vlined]{algorithm2e}

\SetCommentSty{mycommfont}
\SetKwInput{KwInput}{Input}                
\SetKwInput{KwOutput}{Output}  
\SetKwInput{KwParam}{Parameter}

\usepackage[hidelinks]{hyperref}


\title{\LARGE \bf
Robust Single-Parameter Point Cloud Registration \\using Truncated $L_p$ Functions
}

\author{Seong Hun Lee$^{1}$, Patrick Vandewalle$^{2}$ and Javier Civera$^{3}$
\thanks{$^{1,3}$ Seong Hun Lee and Javier Civera are with I3A, University of Zaragoza, 50018 Zaragoza, Spain.
{\tt\small seonghunlee@unizar.es}, {\tt\small jcivera@unizar.es}}%
\thanks{$^{2}$ Patrick Vandewalle is with EAVISE, Department of Electrical Engineering, KU Leuven, 2860 Sint-Katelijne-Waver, Belgium.
        {\tt\small patrick.vandewalle@kuleuven.be}}%
}

\begin{document}

\maketitle
\thispagestyle{empty}
\pagestyle{empty}
\raggedbottom 

\begin{abstract}
We propose a novel method for robust point cloud registration with only one tunable parameter.
Our method, dubbed TLP, uses truncated $L_p$ functions with different values of $p$ to assign priority scores to correspondences, select the hypothesis, and refine the final model.
By leveraging the robustness of truncated functions, TLP effectively handles the high outlier rates commonly encountered in real-world 3D feature matching.
We also exploit GPU parallelization to evaluate hypotheses efficiently, enabling registration from thousands of correspondences in under a second on average.
With only one tunable parameter, our method offers greater practicality than multi-parameter methods that require dataset-specific tuning of several parameters.
Evaluations on public benchmark datasets demonstrate that TLP achieves higher accuracy than state-of-the-art single-parameter methods.
Furthermore, it outperforms two widely used multi-parameter methods on three of the four evaluated datasets while maintaining competitive runtime.
\end{abstract}

\section{INTRODUCTION}

Point cloud registration estimates the rotation and translation needed to align point clouds captured from different viewpoints, enabling 3D reconstruction, localization, and robotic perception~\cite{huang2024tpami,lee2024eccvw}.
A common pipeline first establishes correspondences using local descriptors, such as handcrafted FPFH~\cite{rusu2009icra} or learned FCGF~\cite{choy2019iccv}, and then estimates the transformation.
Correct correspondences allow the least-squares alignment to be computed in closed form~\cite{horn1987absolute}, but partial overlap, ambiguous geometry, and noise can cause outliers to greatly outnumber inliers~\cite{chen2022cvpr,yang2021tro}.
Reliable registration therefore requires identifying geometrically consistent correspondences among sets dominated by outliers.

RANSAC addresses this problem by generating hypotheses from random minimal samples and evaluating their consensus~\cite{fischler1981ransac}, but its efficiency deteriorates as the probability of drawing an all-inlier sample decreases.
Recent methods exploit compatibility graphs, correspondence rankings, and consistency filtering to make hypothesis generation more selective~\cite{chen2022cvpr,zhang2023cvpr,lee2024eccvw,shi2024ral}.
Others combine robust objectives with specialized optimization or parameter search~\cite{yang2021tro,huang2024cvpr,huang2024tpami}.
Together, these advances motivate a coordinated approach to correspondence scoring, hypothesis selection, and refinement.

Coordinating these stages requires accounting for their different objectives: correspondence scoring should preserve degrees of geometric agreement, hypothesis selection should reward a large consensus, and refinement should reduce alignment errors while limiting outlier influence.
In practice, adapting such a pipeline to a new data domain can require tuning multiple interacting parameters, increasing development effort and complicating deployment.

In this work, we propose TLP, a novel method that uses a common family of truncated $L_p$ costs and complementary scores for correspondence ranking, hypothesis selection, and refinement while requiring users to tune a single parameter.
For correspondence ranking, it combines the truncated $L_3$ compatibility score with the support aggregation introduced in HERE~\cite{huang2024tpami}, preserving degrees of pairwise geometric agreement before aggregation.
Hypothesis selection uses the limiting score as $p\to\infty$, equivalent to maximizing the inlier count, while refinement successively uses the costs corresponding to $p=2$, the logarithmic endpoint $p\to0^+$, and $p=0.4$.
With fixed, stage-specific exponents and truncation thresholds derived from a single distance parameter, this formulation accommodates the different objectives of each stage in a unified manner.

\begin{samepage}
Our main contributions are as follows:
\begin{itemize}[leftmargin=*]
    \item A robust point cloud registration framework that uses
    truncated $L_p$ functions with different exponents for
    correspondence scoring, hypothesis selection, and refinement.
    \item A correspondence scoring scheme that incorporates
    continuous truncated $L_3$ compatibility into support
    aggregation, preserving the degree of pairwise agreement.
    \item An efficient implementation combining deterministic
    hypothesis generation, parallel batch evaluation, and staged
    refinement, with only one user-tuned parameter.
\end{itemize}
Our code is publicly available at 
{\url{https://github.com/sunghoon031/TLP}}.
\end{samepage}

\section{RELATED WORK}

\subsection{Compatibility-Based Correspondence Scoring}
Pairwise distance consistency allows us to assess correspondence correctness without first estimating the rigid body transformation.
ROBIN uses such invariants to construct compatibility graphs and reject outliers through graph pruning~\cite{shi2021robin}. 
CLIPPER retains graded consistency in a weighted graph and seeks a dense, mutually consistent subset~\cite{lusk2021clipper}. 
SC$^2$-PCR strengthens pairwise evidence through second-order spatial compatibility~\cite{chen2022cvpr}, whereas MAC extracts multiple maximal cliques to generate transformation hypotheses~\cite{zhang2023cvpr}. 
HERE ranks correspondences by aggregating the support of their compatible neighbors~\cite{huang2024tpami}. In our method, we adopt this aggregation with continuous truncated $L_3$ compatibility, retaining the degree of pairwise agreement when computing correspondence priorities.

Learned estimators also exploit relations among matches. 
DGR predicts correspondence confidence using a 6D convolutional network~\cite{choy2020dgr}, while PointDSC incorporates spatial consistency into learned outlier rejection~\cite{bai2021pointdsc}. 
Our method requires no estimator training and accepts both handcrafted and learned correspondences.

\subsection{Hypothesis Generation and Consensus Selection}
RANSAC estimates models from random minimal samples and selects them by consensus~\cite{fischler1981ransac}. 
Subsequent methods improve different parts of this process.
PROSAC exploits correspondence rankings by progressively enlarging the pool from which samples are drawn~\cite{chum2005prosac}. 
LO-RANSAC refines promising hypotheses through local optimization~\cite{chum2003loransac}, and Graph-Cut RANSAC incorporates spatial coherence into this refinement through graph-cut-based inlier labeling~\cite{barath2018gcransac}. 
MAGSAC marginalizes over a range of noise scales to score and refine models, reducing dependence on a single inlier threshold~\cite{barath2019magsac}.

For point cloud registration, PCR-99 uses ranked correspondences to guide deterministic three-point hypothesis generation~\cite{lee2024eccvw}. 
TCF improves the input to three-point RANSAC through successive length- and angle-based consistency filtering~\cite{shi2024ral}, while HERE uses selected correspondences to guide a decomposed parameter search~\cite{huang2024tpami}. 
Our method combines ranking-guided generation with GPU batch evaluation, inlier-count selection, and local refinement.

\subsection{Robust Objectives and Refinement}
Robust estimation controls the influence of incorrect correspondences through the residual objective. 
FGR uses a Geman--McClure penalty and graduated non-convexity to optimize alignment~\cite{zhou2016fgr}. More generally, graduated non-convexity couples robust costs with non-minimal solvers by progressively introducing non-convexity~\cite{yang2020gnc}.
TEASER uses truncated least squares~\cite{yang2021tro}, whereas TEAR minimizes truncated entry-wise absolute residuals through a decomposed search~\cite{huang2024cvpr}. 

Parameterized loss families, such as Barron's~\cite{barron2019robust}, provide a complementary approach to robust estimation by continuously adjusting the shape of the penalty.
IRLS optimizes robust objectives by solving a sequence of weighted least-squares problems, and its convergence has been studied in the context of non-smooth robust regression~\cite{peng_2022_nips}.
Our method unifies correspondence scoring, hypothesis selection, and refinement through a normalized, truncated, smoothed $L_p$ family with fixed, stage-specific exponents.

\section{PRELIMINARIES}
\subsection{Problem Definition}
Let $\{(\mathbf{a}_i,\mathbf{b}_i)\}_{i=1}^{n}$ be a set of putative correspondences between two 3D point clouds.
The inlier correspondences are related by a rigid body transformation up to noise:
\begin{equation}
    \mathbf{R}\mathbf{a}_i+\mathbf{t}+\mathbf{e}_i=\mathbf{b}_i
    \quad \text{for} \quad i\in\mathcal{I},
\end{equation}
where $\mathbf{R}\in\mathrm{SO}(3)$ and $\mathbf{t}\in\mathbb{R}^3$ denote the rotation and translation, respectively, $\mathbf{e}_i\in\mathbb{R}^3$ represents noise, and $\mathcal{I}$ is the set of inlier indices.
The remaining correspondences are considered outliers and need not satisfy this relation.
The robust point cloud registration problem addressed in this work is to estimate $(\mathbf{R},\mathbf{t})$ from the given correspondences when the inlier set $\mathcal{I}$ is unknown.

\subsection{Normalized, Truncated, Smoothed \texorpdfstring{$L_p$}{Lp} Cost}
\label{subsec:tlp_cost}
For a residual $r\geq0$, truncation threshold $\delta>0$, and a very small smoothing parameter $\epsilon>0$ (which is fixed to $10^{-12}$ in this work and omitted from the function's notation), we define the normalized, truncated, smoothed $L_p$ cost as
\begin{equation}
\label{eq:tlp_cost_score}
 \rho(r;p,\delta)=\min\left\{
 \frac{(r^2+\epsilon)^{p/2}-\epsilon^{p/2}}
 {(\delta^2+\epsilon)^{p/2}-\epsilon^{p/2}},1\right\},
 \quad 0<p<\infty.
\end{equation}
This cost is nondecreasing in $r$, equals zero at $r=0$, and saturates at one for $r\geq\delta$. 
Its untruncated component is a smoothed, shifted, and scaled version of the $p$th power of the residual magnitude.
We define the endpoint costs by the following pointwise limits:
\begin{equation}
\label{eq:tlp_zero}
\rho(r;0,\delta)
:= \lim_{p\to0^+}\rho(r;p,\delta)
= \min\left\{
\frac{\log(1+r^2/\epsilon)}
     {\log(1+\delta^2/\epsilon)},1
\right\},
\end{equation}
\begin{equation}
\label{eq:tlp_infinity}
\rho(r;\infty,\delta)
:= \lim_{p\to\infty}\rho(r;p,\delta)
= \begin{cases}
0, & r<\delta,\\
1, & r\geq\delta.
\end{cases}
\end{equation}
The derivations are provided in Appendix.
Note that the zero endpoint is logarithmic for fixed $\epsilon>0$, rather than a count of nonzero residuals. 
For finite $p>0$, letting $\epsilon\to0^+$ leads to $\min\{(r/\delta)^p,1\}$.

\begin{figure}[!tbp]
 \centering
 \includegraphics[width=0.4\textwidth]{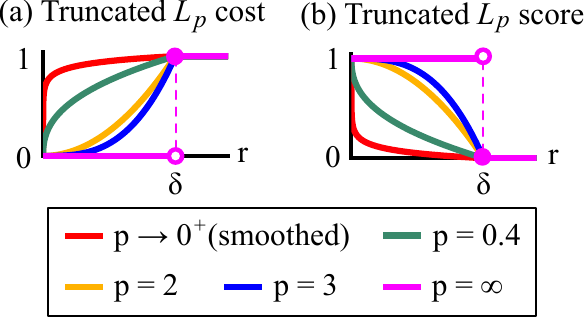}
\caption{Normalized, truncated, smoothed $L_p$ costs and scores versus residual $r$. The truncation threshold is denoted by $\delta$. }
\label{fig:truncated_lp_shapes}
\end{figure}

The complementary score of this cost is $S(r;p,\delta)=1-\rho(r;p,\delta)$. 
For any fixed set of $n$ residuals, maximizing the sum of scores is equivalent to minimizing the sum of costs, since their sums add to $n$. In particular, note that
\begin{equation}
 \sum_{i=1}^{n}S(r_i;\infty,\delta)
 =|\{i:r_i<\delta\}|.
 \label{eq:tlp_infinity_score}
\end{equation}
Thus, maximizing this total score is equivalent to maximizing the number of inliers.
Figure~\ref{fig:truncated_lp_shapes} illustrates both functions.

\section{METHOD}
\subsection{Overview of TLP}
Alg.~\ref{alg:TLP} summarizes the pipeline. TLP first ranks correspondences using truncated $L_3$ compatibility and support aggregation. These rankings guide deterministic three-point hypothesis generation, and GPU batch evaluation selects the two hypotheses with the largest consensus ($p\to\infty$). Each undergoes $L_2$ refinement followed by smoothed $L_{0^+}$ refinement via iteratively reweighted least squares (IRLS)~\cite{peng_2022_nips}\cite[Chapter~9]{lee2023thesis}, with inlier sets updated between model fits. Batch processing continues until the stopping criterion is met, and the best model receives final $L_{0.4}$ refinement. The distance threshold $\tau$ is the only dataset-dependent parameter; all other settings remain fixed.

\begin{figure}[!tbp]
 \centering
 \includegraphics[width=0.4\textwidth]{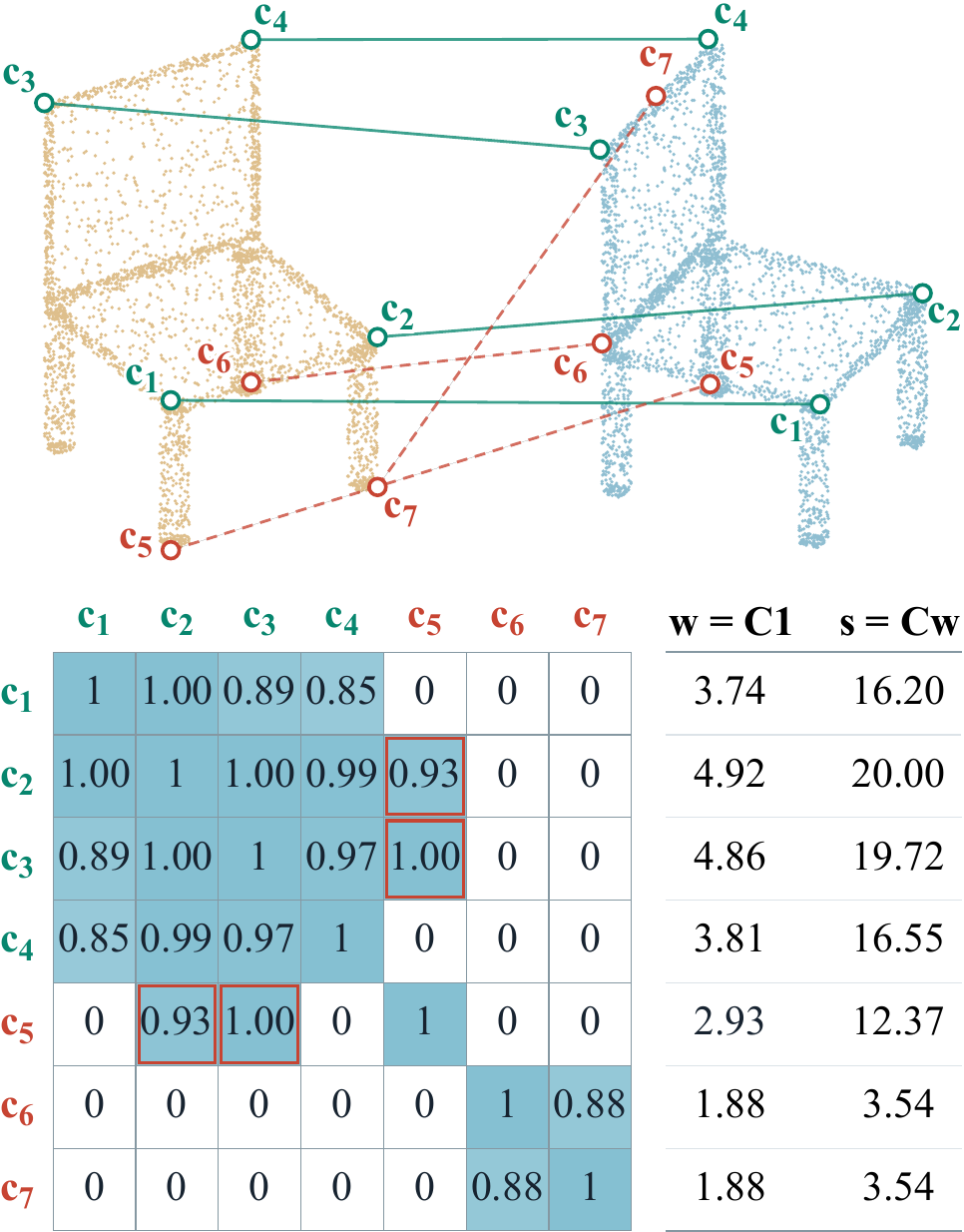}
\caption{\textbf{[Top]} Correct (solid green) and incorrect (dashed red) correspondences. \textbf{[Bottom left]} Truncated $L_3$ compatibility matrix (Sec.~\ref{subsec:correspondence_scoring}). Shaded cells indicate stronger agreement. Red outlines mark accidental compatibilities of $c_5$ with $c_2$ and $c_3$. \textbf{[Bottom right]} Initial support $\mathbf{w}=\mathbf{C}\mathbf{1}$ and aggregated scores $\mathbf{s}=\mathbf{C}\mathbf{w}$. Both rank all correct matches above the incorrect ones. Aggregation increases the ratio of mean inlier-to-outlier ratio from $1.9$ to $2.28$, improving their relative separation in this example.}
\label{fig:tlp_compatibility}
\end{figure}

\begin{algorithm}[!tbp]
\setstretch{0.985}
\caption{TLP}
\label{alg:TLP}
\KwInput{Corresponding 3D point sets
$\{\mathbf{a}_i\}_{i=1}^{n}$ and
$\{\mathbf{b}_i\}_{i=1}^{n}$,
inlier distance threshold $\tau$.}
\KwOutput{Rigid body transformation $(\mathbf{R},\mathbf{t})$.}

\tcc{Construct compatibility matrices and correspondence orderings}
$d_{ij} \gets
\left|
\|\mathbf{a}_i-\mathbf{a}_j\|_2
-
\|\mathbf{b}_i-\mathbf{b}_j\|_2
\right|$
for all $(i,j)$\;

\ForEach{$q$ in $(0.1,\,0.2,\,0.3,\,0.4)$}{
    $\sigma_q \gets 1.2q\tau$\;
    $C^{(q)}_{ij} \gets
    \max\left\{1-\left(d_{ij}/\sigma_q\right)^3,\,0\right\}$
    for all $(i,j)$\;
    $\mathbf{w}^{(q)} \gets \mathbf{C}^{(q)}\mathbf{1}$\;
    $\mathbf{s}^{(q)} \gets \mathbf{C}^{(q)}\mathbf{w}^{(q)}$\;
    $\pi_q \gets$ point indices sorted by descending
    $\mathbf{s}^{(q)}$, with ranks indexed from $0$\;
}

\tcc{Generate and evaluate batches of 256 hypotheses}
$N_{\mathrm{sufficient}}
    \gets \max\{10,\,\lfloor0.05n\rfloor\}$\;
$N_{\mathrm{acceptable}}
    \gets \max\{10,\,\lfloor0.001n\rfloor\}$\;
$N_{\mathrm{best}} \gets 0$, $c \gets 0$, $\mathcal{B} \gets \emptyset$, $\mathcal{S} \gets \emptyset$\;

\For{$h \gets 3$ \KwTo $\min\{1000,\,3n-6\}$}{
    $i_{\min} \gets \max\{0,\,h-(n-2)-(n-1)\}$\;
    $i_{\max} \gets \lfloor(h-3)/3\rfloor$\;

    \For{$i \gets i_{\min}$ \KwTo $i_{\max}$}{
        $j_{\min} \gets \max\{i+1,\,h-i-(n-1)\}$\;
        $j_{\max} \gets \lfloor(h-i-1)/2\rfloor$\;

        \For{$j \gets j_{\min}$ \KwTo $j_{\max}$}{
            $k \gets h-i-j$\;

            \If{$N_{\mathrm{best}} \geq N_{\mathrm{sufficient}}$
                \textbf{or}
                $(c \geq 1500
                \text{ and }
                N_{\mathrm{best}} \geq N_{\mathrm{acceptable}})$}{
                Exit all sampling loops\;
            }

            \ForEach{$q$ in $(0.1,\,0.2,\,0.3,\,0.4)$}{

                \If{$(\pi_q(i),\,\pi_q(j),\,\pi_q(k)) \in \mathcal{S}$}{
                    \textbf{continue}\;
                }
                Add $(\pi_q(i),\,\pi_q(j),\,\pi_q(k))$ to $\mathcal{S}$\;

                Obtain $(\mathbf{R},\mathbf{t})$ by aligning
                the corresponding 3-point sets indexed by $(\pi_q(i),\,\pi_q(j),\,\pi_q(k))$\;
                Add $(\mathbf{R},\mathbf{t})$ to $\mathcal{B}$\;

                \If{$|\mathcal{B}|=256$}{
                    Evaluate $\mathcal{B}$
                    (Alg.~\ref{alg:batch}), updating the best
                    hypothesis and its refined inlier count
                    $N_{\mathrm{best}}$\;
                    $\mathcal{B} \gets \emptyset$\;
                }
            }

            $c \gets c+1$\;
        }
    }
}

\If{$\mathcal{B}\neq\emptyset$}{
    Evaluate $\mathcal{B}$
    (Alg.~\ref{alg:batch}), updating the best hypothesis
    and $N_{\mathrm{best}}$\;
}

$(\mathbf{R},\mathbf{t}) \gets$
Refine the best hypothesis by minimizing a truncated
$L_{0.4}$ cost with truncation threshold $\tau$,
performing at most five inlier set updates
(Alg.~\ref{alg:truncated_lp_minimization})\;

\Return{$(\mathbf{R},\mathbf{t})$}\;
\end{algorithm}

\begin{algorithm}[!tbp]
\label{alg:batch}
\setstretch{1.0}
\caption{Batched hypothesis evaluation}
\KwInput{Corresponding 3D point sets $\{\mathbf{a}_i\}_{i=1}^{n}$ and $\{\mathbf{b}_i\}_{i=1}^{n}$, Inlier distance threshold $\tau$, Batch of hypotheses $\mathcal{B}$.}
\KwOutput{Rigid body transformation $(\mathbf{R}, \mathbf{t})$.}

Evaluate each hypothesis in the batch using the full set of correspondences and compute its inlier count\;

Refine the two hypotheses with the highest inlier counts by minimizing a truncated $L_{2}$ cost with truncation threshold $1.2\tau$, performing at most 50 inlier set updates (Alg.~\ref{alg:truncated_lp_minimization})\;

Refine them again by minimizing a truncated $L_{0^+}$ cost with truncation threshold $1.2\tau$, performing a single inlier set update (Alg.~\ref{alg:truncated_lp_minimization})\;

$(\mathbf{R}, \mathbf{t})\gets$ Select the hypothesis with the higher inlier count\;

\Return{$(\mathbf{R}, \mathbf{t})$;}
\end{algorithm}

\begin{algorithm}[!tbp]
\label{alg:truncated_lp_minimization}
\setstretch{1.0}
\caption{Truncated $L_p$ cost minimization}
\DontPrintSemicolon
\KwInput{Corresponding 3D point sets $\{\mathbf{a}_i\}_{i=1}^{n}$ and $\{\mathbf{b}_i\}_{i=1}^{n}$, Batch of hypotheses $\mathcal{B}$, Initial estimate $(\mathbf{R}_0, \mathbf{t}_0)$, Initial inlier set $\mathcal{I}_0$, Truncation threshold $\delta$, Maximum number of inlier set updates $j_\mathrm{max}$, Parameter $p$ of the $L_p$ norm}
\KwOutput{Rigid body transformation $(\mathbf{R}, \mathbf{t})$, Inlier set $\mathcal{I}$}

$\mathcal{I} \gets \mathcal{I}_0$, $\epsilon \gets 10^{-12}$

\SetKwFunction{Fit}{Fit}
\SetKwProg{Fn}{Function}{:}{}

\Fn{\Fit{$\mathcal{I}$}}{
    \If{$p=2$}{
        \Return{$\displaystyle
        \arg\min_{\mathbf{R}\in\mathrm{SO}(3),\,\mathbf{t}}
        \sum_{i\in\mathcal{I}}
        \|\mathbf{R}\mathbf{a}_i+\mathbf{t}-\mathbf{b}_i\|_2^2$}\;
    }

    $w_i \gets 1$ for all $i\in\mathcal{I}$\;

    \For{$k \gets 1$ \KwTo $10$}{
        $(\mathbf{R},\mathbf{t}) \gets
        \displaystyle\arg\min_{\mathbf{R}\in\mathrm{SO}(3),\,\mathbf{t}}
        \sum_{i\in\mathcal{I}}
        w_i\|\mathbf{R}\mathbf{a}_i+\mathbf{t}-\mathbf{b}_i\|_2^2$\;

        $\widetilde{w}_i \gets
        \left(
        \|\mathbf{R}\mathbf{a}_i+\mathbf{t}-\mathbf{b}_i\|_2^2
        +\epsilon
        \right)^{p/2-1}$
        for all $i\in\mathcal{I}$\;

        $\gamma \gets
        \|\widetilde{\mathbf{w}}-\mathbf{w}\|_2/
        (\|\mathbf{w}\|_2+\epsilon)$\;
        $\mathbf{w} \gets \widetilde{\mathbf{w}}$\;

        \If{$\gamma < 0.03$}{
            \textbf{break}\;
        }
    }

    \Return{$\displaystyle
    \arg\min_{\mathbf{R}\in\mathrm{SO}(3),\,\mathbf{t}}
    \sum_{i\in\mathcal{I}}
    w_i\|\mathbf{R}\mathbf{a}_i+\mathbf{t}-\mathbf{b}_i\|_2^2$}\;
}

\For{$j \gets 1$ \KwTo $j_\mathrm{max}$}{
    $(\mathbf{R},\mathbf{t}) \gets$ \Fit{$\mathcal{I}$}\;

    $\mathcal{I}' \gets
    \left\{
    i \in \{1,\ldots,n\} :
    \|\mathbf{R}\mathbf{a}_i+\mathbf{t}-\mathbf{b}_i\|_2^2
    < \delta^2
    \right\}$\;

    \If{$|\mathcal{I}'|<3$ \textbf{or} $\mathcal{I}'=\mathcal{I}$}{
        \Return{$(\mathbf{R},\mathbf{t})$ and $\mathcal{I}$}\;
    }

    $\mathcal{I} \gets \mathcal{I}'$\;
}

$(\mathbf{R},\mathbf{t}) \gets$ \Fit{$\mathcal{I}$}\;

\Return{$(\mathbf{R}, \mathbf{t})$ and $\mathcal{I}$;}
\end{algorithm}

\subsection{Correspondence Scoring}
\label{subsec:correspondence_scoring}

We score each point correspondence based on how likely it is to be an inlier.
Since a rigid transformation preserves pairwise distances, we first measure the inconsistency between correspondences $i$ and $j$ by
\begin{equation}
 d_{ij}=\left|\|\mathbf{a}_i-\mathbf{a}_j\|_2
              -\|\mathbf{b}_i-\mathbf{b}_j\|_2\right|.
\end{equation}
Then, we construct $n\times n$ truncated $L_3$ compatibility matrices in the following form:
\begin{equation}
    C_{ij}^{(q)}=\max\left\{1-\left(\frac{d_{ij}}{\sigma_q}\right)^3, \ 0\right\} \quad \text{with} \quad \sigma_q=1.2q\tau,
\end{equation}
where $q\in\{0.1,\ 0.2,\ 0.3,\ 0.4\}$.
Thus, correspondences with identical pairwise distances receive compatibility one, while discrepancies at or above $\sigma_q$ receive zero.
Intermediate discrepancies receive continuous weights rather than binary classifications, as used in \cite{chen2022cvpr, huang2024tpami}.

We adopt the scoring approach of \cite{huang2024tpami}, replacing its binary compatibility matrix with our continuous one.
For each $q\in\{0.1,\ 0.2,\ 0.3,\ 0.4\}$, we compute 
\begin{equation}
 \mathbf{w}^{(q)}=\mathbf{C}^{(q)}\mathbf{1},\qquad
 \mathbf{s}^{(q)}=\mathbf{C}^{(q)}\mathbf{w}^{(q)}.
 \label{eq:correspondence_scores}
\end{equation}
The first multiplication measures each correspondence's total compatibility. 
The second weights its compatible neighbors by their own support, favoring correspondences that agree with other well-supported correspondences.

Figure~\ref{fig:tlp_compatibility} illustrates how this two-stage scoring scheme ranks correct correspondences above incorrect ones, even when some incorrect correspondences exhibit high pairwise compatibility.
Sorting $\mathbf{s}^{(q)}$ in descending order produces a permutation $\pi_q$ of the correspondence indices.
The four fixed values of $q$ provide four orderings, allowing hypothesis generation to explore different compatibility neighborhoods without independently tuning four thresholds.

\subsection{Batched Hypothesis Generation}

We generate three-point samples deterministically from the score orderings.
Each permutation $\pi_q$ maps ranks $0,\ldots,n-1$ to the $n$ point correspondence using the method proposed in \cite{lee2024eccvw}.
This method enumerates triplets $(i,j,k)$ satisfying $0 \leq i < j < k < n$ in increasing order of their index sum $s=i+j+k$.
For each sum, $i$ and then $j$ are traversed in ascending order.
The enumeration begins as follows:
\begin{itemize}
    \item All triplets with sum 3: $(0, 1, 2)$.
    \item All triplets with sum 4: $(0, 1, 3)$.
    \item All triplets with sum 5: $(0, 1, 4)$, $(0,2,3)$.
    \item All triplets with sum 6: $(0, 1, 5)$, $(0,2,4)$, $(1, 2, 3)$.
    \item ...
    \item Unless an early termination condition is met, enumeration continues up to an index sum of $\min(1000,3n-6)$, reaching the final triplet $(n-3,n-2,n-1)$ when $3n-6\leq1000$.
\end{itemize}
This visits combinations of highly ranked correspondences first. 
For each rank triplet and each $q\in\{0.1,\ 0.2,\ 0.3,\ 0.4\}$, we form the correspondence sample $(\pi_q(i),\pi_q(j),\pi_q(k))$.
Samples containing the same three original correspondence indices are evaluated only once, even if they arise from different orderings. 
We collect 256 distinct samples per batch and estimate and evaluate their transformations in parallel.

For each sample, a relative transformation is obtained by aligning local frames constructed from the ordered source and target triplets. 
Given the ordered sample indices $(i_1,i_2,i_3)$, the source frame is defined as
\begin{align}
 \mathbf{u}&=\mathbf{a}_{i_2}-\mathbf{a}_{i_1},
 &\mathbf{v}&=\mathbf{a}_{i_3}-\mathbf{a}_{i_1},\\
 \mathbf{f}_1&=\nu(\mathbf{u}),
 &\mathbf{f}_2&=\nu(\mathbf{u}\times\mathbf{v}),\\
 \mathbf{f}_3&=\nu(\mathbf{f}_1\times\mathbf{f}_2),
 &\mathbf{F}_a&=[\mathbf{f}_1\ \mathbf{f}_2\ \mathbf{f}_3],
\end{align}
where $\nu(\mathbf{z})=\mathbf{z}/\sqrt{\max(\|\mathbf{z}\|_2^2,\epsilon)}$ with $\epsilon=10^{-12}$ provides numerically safeguarded normalization. The target frame $\mathbf{F}_b$ is constructed analogously from $(\mathbf{b}_{i_1},\mathbf{b}_{i_2},\mathbf{b}_{i_3})$. 
The rotation and translation are then
\begin{equation}
 \mathbf{R}_h=\mathbf{F}_b\mathbf{F}_a^{\mathsf T},\qquad
 \mathbf{t}_h=\bar{\mathbf{b}}_h-\mathbf{R}_h\bar{\mathbf{a}}_h,
 \label{eq:triplet_transform}
\end{equation}
where $\bar{\mathbf{a}}_h$ and $\bar{\mathbf{b}}_h$ denote the source and target triplet centroids, respectively.
For nondegenerate triplets, the frames are orthonormal and yield a rigid transformation.

After processing each batch using the selection and refinement procedure below, we retain the largest consensus found across all batches. 
Let $n^\star$ be its size and let $m$ count the rank triplets visited. 
We define
\begin{equation}
 \begin{aligned}
 n_{\mathrm{suff}}&=\max\{10,\lfloor0.05n\rfloor\},\\
 n_{\mathrm{acc}}&=\max\{10,\lfloor0.001n\rfloor\}.
 \end{aligned}
\end{equation}

Generation stops when $n^\star\geq n_{\mathrm{suff}}$, or when $m\geq1500$ and $n^\star\geq n_{\mathrm{acc}}$. 
Any remaining partial batch is processed before final refinement.

\subsection{Truncated \texorpdfstring{$L_p$}{Lp} Cost Minimization}

Let $r_i(\mathbf{R},\mathbf{t})=\|\mathbf{R}\mathbf{a}_i+\mathbf{t}-\mathbf{b}_i\|_2$. 
Using the cost defined in Section~\ref{subsec:tlp_cost}, the proposed refinement seeks to minimize
\begin{equation}
 \min_{\mathbf{R}\in\mathrm{SO}(3),\,\mathbf{t}}
 \sum_{i=1}^{n}\rho(r_i(\mathbf{R},\mathbf{t});p,\delta).
 \label{eq:tlp_refinement_objective}
\end{equation}
To this end, we alternate model fitting on the current inlier set $\mathcal{I}$ and thresholding all residuals to update $\mathcal{I}$.
Residuals above the threshold incur a constant cost of one, so model fitting considers only the untruncated cost of the current inliers.

For fixed $p$ and $\delta$, the normalization and additive offset in~\eqref{eq:tlp_cost_score} do not affect the fitted model. 
Differentiating the untruncated cost with respect to the squared residual and dropping the common factor, we get the following IRLS weights:
\begin{equation}
 w_i=(r_i^2+\epsilon)^{p/2-1},\qquad i\in\mathcal{I}.
 \label{eq:tlp_irls_weights}
\end{equation}
Each IRLS step solves a weighted rigid least-squares problem and recomputes the weights from the resulting residuals. For $p=2$, the weights are constant and a single least-squares fit suffices. 

Algorithm~\ref{alg:truncated_lp_minimization} details our implementation.
We use $p=2$ followed by $p=0^+$ with $\delta=1.2\tau$ during batch evaluation (see Alg.~\ref{alg:batch}), and $p=0.4$ with $\delta=\tau$ for final refinement (see Alg.~\ref{alg:TLP}).

\section{RESULTS}
\label{sec:results}
\subsection{Experimental Setup}
We compare our method, TLP, against five state-of-the-art single-parameter methods (TEASER++~\cite{yang2021tro}\footnote{We use the C++ version: \url{https://github.com/MIT-SPARK/TEASER-plusplus}}, TEAR~\cite{huang2024cvpr}\footnote{\url{https://github.com/tyhuang98/TEAR-release}}, HERE~\cite{huang2024tpami}\footnote{\url{https://github.com/tyhuang98/HERE-release}}, PCR-99~\cite{lee2024eccvw}\footnote{We use the Python version (with GPU parallelization): \url{https://github.com/sunghoon031/PCR-99-python}} and TCF~\cite{shi2024ral}\footnote{\url{https://github.com/ShiPC-AI/TCF}}) and two widely used multi-parameter methods (SC$^2$-PCR~\cite{chen2022cvpr}\footnote{\url{https://github.com/ZhiChen902/SC2-PCR}}  and MAC~\cite{zhang2023cvpr}\footnote{We use the C++ version in Linux folder: \url{https://github.com/zhangxy0517/3d-registration-with-maximal-cliques}}).
For HERE~\cite{huang2024tpami}, we evaluate two variants: with and without global scaling of the point clouds.
Among the evaluated methods, TLP, SC$^2$-PCR, and PCR-99 are implemented in Python, while the remaining methods are implemented in C++.
All experiments are conducted on a laptop equipped with an Intel Core i7-7700HQ CPU, an NVIDIA Quadro M1200 GPU, and 16 GB of RAM.

We evaluate all methods on the 3DMatch~\cite{zeng2017cvpr} and KITTI~\cite{geiger2012cvpr} datasets using FPFH~\cite{rusu2009icra} and FCGF~\cite{choy2019iccv} descriptors.
We obtain the 3D--3D correspondence data using the same processing pipeline as SC$^2$-PCR~\cite{chen2022cvpr}.
As a result, the 3DMatch and KITTI datasets contain 1,623 and 555 point-cloud pairs, respectively, for each descriptor setting.
Registration is performed on these precomputed correspondences.

For the two multi-parameter methods, SC$^2$-PCR~\cite{chen2022cvpr} and MAC~\cite{zhang2023cvpr}, we use the parameter settings recommended by their authors.
For the single-parameter methods, we tune the thresholds separately for each dataset--descriptor configuration by maximizing mAA at $15^\circ$ (see Sec.~\ref{subsec:metric} for the definition of this metric). 
Table~\ref{tab:thresholds} lists the threshold values we use.

\begin{table}[!tbp]
\centering
\caption{Selected distance thresholds for each dataset--descriptor configuration, tuned by maximizing mAA at $15^\circ$. 
N/A denotes an unavailable setting.}
\label{tab:thresholds}
\small
\renewcommand{\arraystretch}{1.2}
\begin{tabular}{lcccc}
\toprule
& \multicolumn{2}{c}{3DMatch} & \multicolumn{2}{c}{KITTI} \\
\cmidrule(lr){2-3}\cmidrule(lr){4-5}
Method & FPFH & FCGF & FPFH & FCGF \\
\midrule
TEASER++             & 0.04 & N/A  & 0.45 & N/A \\
TEAR               & 0.07 & 0.07 & 0.4  & 2.1 \\
HERE (w/o scaling) & 0.07 & 0.07 & 0.7  & 1.8 \\
HERE (with scaling)& 0.07 & 0.07 & 0.7  & 1.7 \\
PCR-99             & 0.02 & 0.01 & 0.25  & 0.4 \\
TCF                & 0.045 & 0.03 & 0.3  & 1.2 \\
TLP (Ours)         & 0.09 & 0.07 & 0.9  & 1.8 \\
\bottomrule
\end{tabular}
\end{table}

\begin{figure*}[!tbp]
 \centering
 \includegraphics[width=\textwidth]{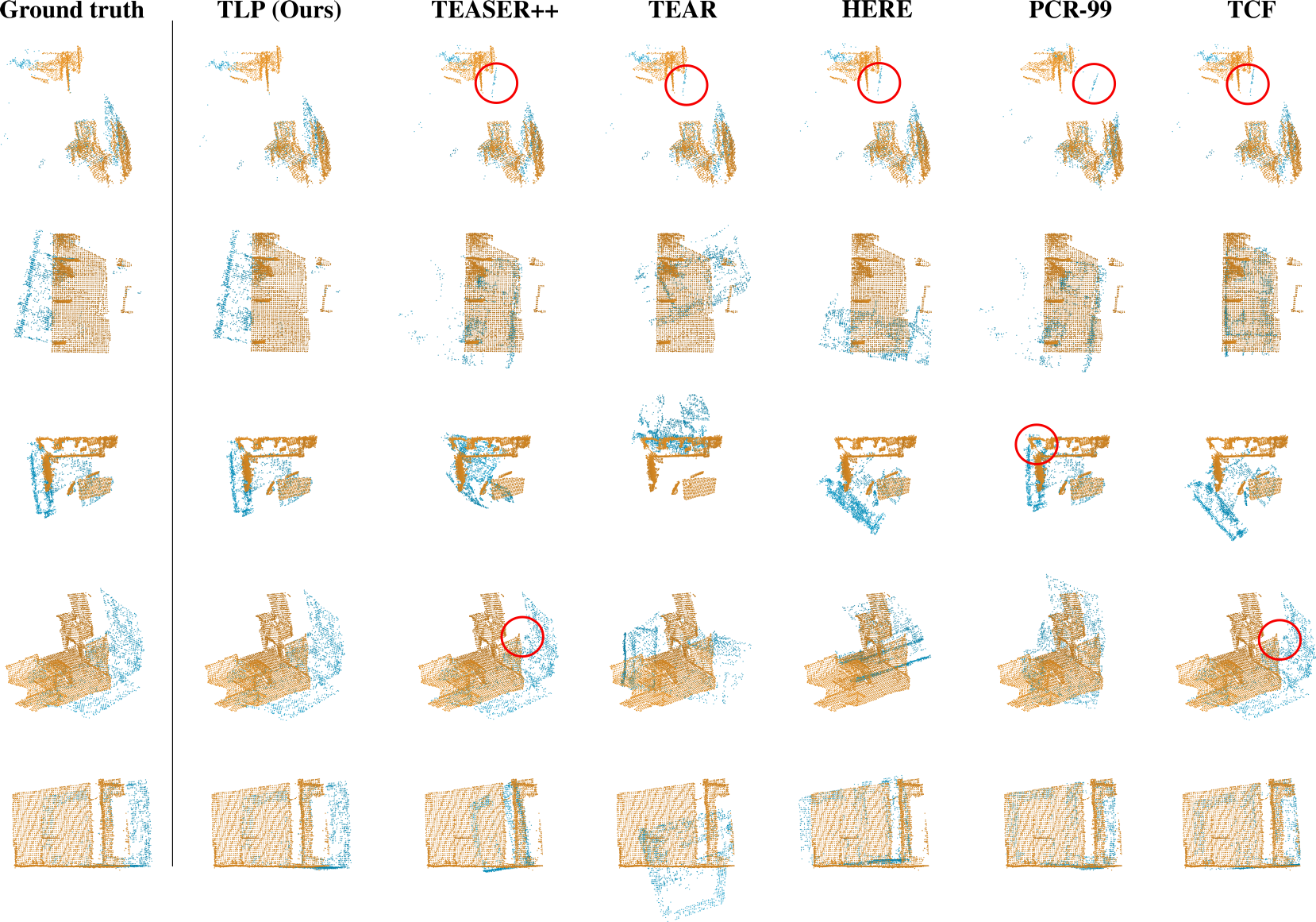}
\caption{Qualitative comparison on five point-cloud pairs from 3DMatch using FPFH descriptors. The two point clouds are shown in blue and orange. TLP produces alignments closely matching the ground truth, while the single-parameter baselines exhibit varying degrees of misalignment. Red circles highlight selected alignment errors.}
\label{fig:qualitative_3dmatch}
\end{figure*}

\begin{table*}[!tbp]
\centering
\caption{Rotation-based registration accuracy (mAA) across different datasets and features. Higher values indicate better performance, with a maximum of 1. The best available result in each row is highlighted in blue and bold.}
\label{tab:registration_accuracy}
\resizebox{\textwidth}{!}{%
\renewcommand{\arraystretch}{1.2}
\begin{tabular}{ll*{9}{c}}
\toprule
& & \multicolumn{2}{c}{Multi-Parameter}
  & \multicolumn{7}{c}{Single-Parameter} \\
\cmidrule(lr){3-4}\cmidrule(lr){5-11}
Dataset & Metric
& SC$^2$-PCR & MAC & TEASER++ & TEAR
& \makecell[cc]{HERE\\(with scaling)}
& \makecell[cc]{HERE\\(w/o scaling)}
& PCR-99 & TCF & \makecell[cc]{TLP\\(Ours)} \\
\midrule

\multirow[c]{3}{*}{\makecell[cc]{3DMatch\\+ FPFH}}
& mAA at $5^\circ$
& 0.5861
& \multirow[c]{3}{*}{\makecell[cc]{Out of\\memory}}
& 0.5503 & 0.3257 & 0.4731 & 0.4734 & 0.5110 & 0.5540
& {\color{blue}\textbf{0.5879}} \\
& mAA at $10^\circ$
& {\color{blue}\textbf{0.7129}}
&
& 0.6747 & 0.3835 & 0.5587 & 0.5596 & 0.6316 & 0.6808
& 0.7002 \\
& mAA at $15^\circ$
& {\color{blue}\textbf{0.7649}}
&
& 0.7272 & 0.4118 & 0.5968 & 0.5979 & 0.6848 & 0.7321
& 0.7456 \\
\midrule

\multirow[c]{3}{*}{\makecell[cc]{3DMatch\\+ FCGF}}
& mAA at $5^\circ$
& 0.6612 & 0.6662
& \multirow[c]{3}{*}{\makecell[cc]{Excessive\\runtime\\(some $>2$ h)}}
& 0.5569 & 0.6430 & 0.6442 & 0.6237 & 0.6295
& {\color{blue}\textbf{0.6818}} \\
& mAA at $10^\circ$
& 0.7955 & 0.7977
&
& 0.6734 & 0.7622 & 0.7631 & 0.7619 & 0.7619
& {\color{blue}\textbf{0.8048}} \\
& mAA at $15^\circ$
& 0.8468 & 0.8475
&
& 0.7213 & 0.8056 & 0.8070 & 0.8174 & 0.8107
& {\color{blue}\textbf{0.8517}} \\
\midrule

\multirow[c]{3}{*}{\makecell[cc]{KITTI\\+ FPFH}}
& mAA at $5^\circ$
& 0.9870
& \multirow[c]{3}{*}{\makecell[cc]{Out of\\memory}}
& 0.9867 & 0.6908 & 0.9712 & 0.9712 & 0.9521 & 0.9845
& {\color{blue}\textbf{0.9899}} \\
& mAA at $10^\circ$
& 0.9935
&
& 0.9933 & 0.7281 & 0.9800 & 0.9793 & 0.9735 & 0.9923
& {\color{blue}\textbf{0.9950}} \\
& mAA at $15^\circ$
& 0.9957
&
& 0.9956 & 0.7540 & 0.9837 & 0.9826 & 0.9813 & 0.9948
& {\color{blue}\textbf{0.9966}} \\
\midrule

\multirow[c]{3}{*}{\makecell[cc]{KITTI\\+ FCGF}}
& mAA at $5^\circ$
& 0.9859 & 0.9852
& \multirow[c]{3}{*}{\makecell[cc]{Excessive\\runtime\\(often $>0.5$ h)}}
& 0.9744 & 0.9885 & 0.9881 & 0.9874 & 0.9892
& {\color{blue}\textbf{0.9924}} \\
& mAA at $10^\circ$
& 0.9885 & 0.9881
&
& 0.9791 & 0.9924 & 0.9923 & 0.9919 & 0.9928
& {\color{blue}\textbf{0.9944}} \\
& mAA at $15^\circ$
& 0.9893 & 0.9891
&
& 0.9810 & 0.9938 & 0.9936 & 0.9934 & 0.9940
& {\color{blue}\textbf{0.9951}} \\
\bottomrule
\end{tabular}%
}
\end{table*}

\begin{table*}[!tbp]
\centering
\caption{Runtime comparison. The fastest result on each dataset is shown in blue and bold.}
\label{tab:runtime}
\small
\begin{tabular}{l*{9}{c}}
\toprule
& \multicolumn{2}{c}{Multi-Parameter}
& \multicolumn{7}{c}{Single-Parameter} \\
\cmidrule(lr){2-3}\cmidrule(lr){4-10}
Dataset
& SC$^2$-PCR & MAC & TEASER++ & TEAR
& \makecell[cc]{HERE\\(with scaling)}
& \makecell[cc]{HERE\\(w/o scaling)}
& PCR-99 & TCF & \makecell[cc]{TLP\\(Ours)} \\
\midrule

\makecell[cc]{3DMatch\\+ FPFH}
& {\color{blue}\textbf{0.24 s}}
& \makecell[cc]{Out of\\memory}
& 1.45 s & 1.64 s & 1.87 s & 2.21 s & 0.68 s & 0.30 s & 0.36 s \\
\addlinespace

\makecell[cc]{3DMatch\\+ FCGF}
& 0.24 s & 11.46 s
& \makecell[cc]{Sometimes\\$>2$ h}
& 0.99 s & 1.06 s & 1.49 s & 0.22 s & 2.68 s & {\color{blue}\textbf{0.20 s}} \\
\addlinespace

\makecell[cc]{KITTI\\+ FPFH}
& 0.71 s
& \makecell[cc]{Out of\\memory}
& 1.46 s & 3.53 s & 2.02 s & 4.80 s & 0.19 s & {\color{blue}\textbf{0.10 s}} & 0.65 s \\
\addlinespace

\makecell[cc]{KITTI\\+ FCGF}
& 0.73 s & 22.46 s
& \makecell[cc]{Often\\$>0.5$ h}
& 0.29 s & 1.19 s & 3.21 s & {\color{blue}\textbf{0.18 s}} & 39.33 s & 0.40 s \\

\bottomrule
\end{tabular}
\end{table*}

\begin{table*}[!tbp]
\centering
\caption{Ablation study: Each variant removes the indicated setting or refinement step from the default configuration. Performance is measured by rotation-based registration accuracy (mAA). Higher values indicate better performance, with a maximum of 1. The best available result in each row is highlighted in blue and bold.} 
\label{tab:ablation}
\resizebox{\textwidth}{!}{%
\renewcommand{\arraystretch}{1.2}
\begin{tabular}{ll*{9}{c}}
\toprule
Dataset & Metric
& \makecell[cc]{TLP\\(Default)}
& \makecell[cc]{Without $1.2$\\factor for $\tau$}
& \makecell[cc]{Without\\$q=0.1$}
& \makecell[cc]{Without\\$q=0.2$}
& \makecell[cc]{Without\\$q=0.3$}
& \makecell[cc]{Without\\$q=0.4$}
& \makecell[cc]{Without $L_2$\\refinement}
& \makecell[cc]{Without $L_{0^+}$\\refinement}
& \makecell[cc]{Without $L_{0.4}$\\refinement} \\
\midrule

\multirow[c]{3}{*}{\makecell[cc]{3DMatch\\+ FPFH}}
& mAA at $5^\circ$
& {\color{blue}\textbf{0.5879}}
& 0.5813 & 0.5875 & 0.5845 & 0.5864 & 0.5850
& 0.5809 & 0.5774 & 0.5791 \\
& mAA at $10^\circ$
& {\color{blue}\textbf{0.7002}}
& 0.6958 & 0.6962 & 0.6952 & 0.6988 & 0.6966
& 0.6944 & 0.6861 & 0.6948 \\
& mAA at $15^\circ$
& {\color{blue}\textbf{0.7456}}
& 0.7420 & 0.7405 & 0.7397 & 0.7441 & 0.7413
& 0.7401 & 0.7299 & 0.7416 \\
\midrule

\multirow[c]{3}{*}{\makecell[cc]{3DMatch\\+ FCGF}}
& mAA at $5^\circ$
& 0.6818 & 0.6792 & 0.6817 & 0.6819 & 0.6821 & 0.6822
& 0.6812 & {\color{blue}\textbf{0.6826}} & 0.6764 \\
& mAA at $10^\circ$
& 0.8048 & 0.8032 & 0.8047 & 0.8046 & 0.8047 & 0.8044
& 0.8033 & {\color{blue}\textbf{0.8052}} & 0.8017 \\
& mAA at $15^\circ$
& 0.8517 & 0.8506 & 0.8516 & 0.8515 & 0.8516 & 0.8512
& 0.8495 & {\color{blue}\textbf{0.8518}} & 0.8497 \\
\midrule

\multirow[c]{3}{*}{\makecell[cc]{KITTI\\+ FPFH}}
& mAA at $5^\circ$
& {\color{blue}\textbf{0.9899}}
& {\color{blue}\textbf{0.9899}}
& {\color{blue}\textbf{0.9899}}
& 0.9896
& {\color{blue}\textbf{0.9899}}
& {\color{blue}\textbf{0.9899}}
& 0.9896 & 0.9867 & 0.9892 \\
& mAA at $10^\circ$
& {\color{blue}\textbf{0.9950}}
& 0.9941 & 0.9942 & 0.9939
& {\color{blue}\textbf{0.9950}}
& {\color{blue}\textbf{0.9950}}
& 0.9942 & 0.9915 & 0.9946 \\
& mAA at $15^\circ$
& {\color{blue}\textbf{0.9966}}
& 0.9954 & 0.9962 & 0.9953
& {\color{blue}\textbf{0.9966}}
& {\color{blue}\textbf{0.9966}}
& 0.9962 & 0.9932 & 0.9964 \\
\midrule

\multirow[c]{3}{*}{\makecell[cc]{KITTI\\+ FCGF}}
& mAA at $5^\circ$
& {\color{blue}\textbf{0.9924}}
& {\color{blue}\textbf{0.9924}}
& {\color{blue}\textbf{0.9924}}
& {\color{blue}\textbf{0.9924}}
& {\color{blue}\textbf{0.9924}}
& {\color{blue}\textbf{0.9924}}
& {\color{blue}\textbf{0.9924}}
& {\color{blue}\textbf{0.9924}}
& 0.9921 \\
& mAA at $10^\circ$
& {\color{blue}\textbf{0.9944}}
& {\color{blue}\textbf{0.9944}}
& {\color{blue}\textbf{0.9944}}
& {\color{blue}\textbf{0.9944}}
& {\color{blue}\textbf{0.9944}}
& {\color{blue}\textbf{0.9944}}
& {\color{blue}\textbf{0.9944}}
& {\color{blue}\textbf{0.9944}}
& 0.9942 \\
& mAA at $15^\circ$
& {\color{blue}\textbf{0.9951}}
& {\color{blue}\textbf{0.9951}}
& {\color{blue}\textbf{0.9951}}
& {\color{blue}\textbf{0.9951}}
& {\color{blue}\textbf{0.9951}}
& {\color{blue}\textbf{0.9951}}
& {\color{blue}\textbf{0.9951}}
& {\color{blue}\textbf{0.9951}}
& 0.9950 \\
\bottomrule
\end{tabular}%
}
\end{table*}

\subsection{Evaluation Metric}
\label{subsec:metric}
We evaluate point cloud registration accuracy using the mean average accuracy (mAA) of the estimated rotations \cite{nonsac}:
\begin{equation}
\label{eq:mAA}
    \textrm{mAA} \ \text{at} \ K^\circ = \frac{1}{KN_p}\sum_{\theta=1^\circ}^{K^\circ}\sum_{i=1}^{N_{\scaleto{p}{4pt}}}f(\theta, e_i), \quad K\in\{5,10,15\},
\end{equation}
where $N_p$ is the number of point cloud pairs (1,623 for 3DMatch and 555 for KITTI datasets), $e_i$ is the rotation error (in degrees) with respect to the ground truth for pair $i$, and $f(\cdot)$ is the indicator function, \emph{i.e.},
\begin{equation}
    f(\theta, e_i) = 
    \begin{cases}
    1, & \text{if } e_i < \theta \\
    0, & \text{otherwise.}\\
  \end{cases}
\end{equation}
Thus, mAA averages accuracy over integer angular thresholds from $1^\circ$ to $K^\circ$, rewarding smaller errors within this range without penalizing the magnitude of errors beyond it. 
As a result, mAA reflects both estimation accuracy and robustness to outliers while reducing dependence on any single evaluation threshold.
The score lies in $[0,1]$ and is independent of point cloud scale, allowing the same thresholds across datasets. 
The main limitation of rotation-based mAA is that it does not measure translation accuracy: a high score alone does not guarantee correct alignment.
In practice, however, accurate rotation estimates tend to be strongly correlated with accurate translation estimates.
Thus, rotation-based mAA serves as a useful, scale-independent proxy for overall registration quality.
\subsection{Registration Accuracy}
Table~\ref{tab:registration_accuracy} summarizes the registration accuracy results.
Some baseline evaluations could not be completed under our experimental conditions. 
MAC ran out of memory for both FPFH configurations.
TEASER incurred excessive runtimes with FCGF, exceeding two hours for some 3DMatch pairs and frequently exceeding half an hour for KITTI pairs.
These entries are reported as unavailable, and the corresponding TEASER threshold entries are marked N/A in Table~\ref{tab:thresholds}.

Among the available results, TLP achieves the highest mAA in 10 of the 12 dataset--descriptor--cutoff combinations.
It consistently outperforms all evaluated single-parameter baselines.
It also outperforms all evaluated multi-parameter baselines, except on 3DMatch with FPFH at the $10^\circ$ and $15^\circ$ cutoffs, where SC$^2$-PCR achieves higher accuracy.
Fig.~\ref{fig:qualitative_3dmatch} presents five examples from 3DMatch with FPFH, where TLP achieves more accurate alignment than the evaluated single-parameter baselines.

\subsection{Ablation Study}
Table~\ref{tab:ablation} summarizes the results of our ablation study.
It shows that the default configuration achieves the highest or joint-highest mAA in 9 of the 12 settings.
Removing individual components generally reduces accuracy on 3DMatch with FPFH, whereas the effects on KITTI are small. 

\subsection{Runtime}
Table~\ref{tab:runtime} reports the runtimes. 
Among the nine methods, TLP ranks third, first, third, and third in runtime across the four dataset--descriptor combinations, respectively.
Thus, TLP consistently ranks among the three fastest methods while achieving superior accuracy in most evaluated settings.

\section{CONCLUSION}
We presented TLP, a robust point cloud registration method that uses truncated $L_p$ functions with different exponents for correspondence scoring, hypothesis selection, and refinement. By combining continuous compatibility scores with deterministic hypothesis generation, parallel batch evaluation, and staged refinement, TLP requires only one user-tuned distance threshold, with all other settings fixed across datasets. Experiments on 3DMatch and KITTI with FPFH and FCGF descriptors show that TLP achieves the highest rotation-based mAA in 10 of the 12 evaluated settings among the available baseline results, while ranking among the three fastest methods in all four dataset--descriptor configurations. These results demonstrate that the proposed combination provides accurate and efficient registration with limited parameter tuning.

\section*{APPENDIX: Endpoint Limits of the Truncated $L_p$ Cost}

Fix $r\geq0$, $\delta>0$, and $\epsilon>0$, and write
\begin{equation}
 A=1+r^2/\epsilon,\qquad B=1+\delta^2/\epsilon.
\end{equation}
Then $A\geq1$, $B>1$, and division by $\epsilon^{p/2}$ in~\eqref{eq:tlp_cost_score} gives
\begin{equation}
 \rho(r;p,\delta)
 =\min\left\{\frac{A^{p/2}-1}{B^{p/2}-1},1\right\}.
 \label{eq:tlp_limit_ratio}
\end{equation}
As $p\to0^+$, the expansion $e^x=1+x+O(x^2)$ yields
\begin{equation}
 \frac{A^{p/2}-1}{B^{p/2}-1}
 =\frac{(p/2)\log A+O(p^2)}{(p/2)\log B+O(p^2)}
 \longrightarrow\frac{\log A}{\log B}.
\end{equation}
Since $\log B>0$, substituting this into~\eqref{eq:tlp_limit_ratio} proves~\eqref{eq:tlp_zero}.

For $p\to\infty$ and $r<\delta$, we have $A/B<1$ and
\begin{equation}
 \frac{A^{p/2}-1}{B^{p/2}-1}
 =\frac{(A/B)^{p/2}-B^{-p/2}}{1-B^{-p/2}}
 \longrightarrow0.
\end{equation}
If $r=\delta$, the ratio is exactly one for every $p>0$. If $r>\delta$, the ratio exceeds one and truncation makes the cost exactly one. These cases prove~\eqref{eq:tlp_infinity}.

\balance

\end{document}